\documentclass[11pt,a4paper]{article}
\usepackage[hyperref]{emnlp2020}
\usepackage{times}
\usepackage{latexsym}
\usepackage{fancyhdr,graphicx}
\usepackage{multirow}
\usepackage{amsmath,amsfonts,amssymb}
\usepackage{subcaption}
\usepackage{url}

\usepackage{microtype}

\usepackage{array}
\newcolumntype{C}{>{\centering\arraybackslash}p{3em}}

\aclfinalcopy 

\title{Exploring Fine-tuning Techniques for Pre-trained Cross-lingual \\ Models via Continual Learning}

\author{Zihan Liu, Genta Indra Winata, Andrea Madotto, Pascale Fung \\
Center for Artificial Intelligence Research (CAiRE)\\
Department of Electronic and Computer Engineering\\
The Hong Kong University of Science and Technology, Clear Water Bay, Hong Kong\\
\texttt{zihan.liu@connect.ust.hk}}

\date{}

\begin{document}
\maketitle
\begin{abstract}
Recently, fine-tuning pre-trained language models (e.g., multilingual BERT) to downstream cross-lingual tasks has shown promising results.
However, the fine-tuning process inevitably changes the parameters of the pre-trained model and weakens its cross-lingual ability, which leads to sub-optimal performance.
To alleviate this problem, we leverage continual learning to preserve the original cross-lingual ability of the pre-trained model when we fine-tune it to downstream tasks.
The experimental result shows that our fine-tuning methods can better preserve the cross-lingual ability of the pre-trained model in a sentence retrieval task. Our methods also achieve better performance than other fine-tuning baselines on the zero-shot cross-lingual part-of-speech tagging and named entity recognition tasks.

\end{abstract}

\section{Introduction}
Recently, multilingual language models~\cite{devlin2019bert, nipscross2019}, pre-trained on extensive monolingual or bilingual resources across numerous languages, have been shown to enjoy surprising cross-lingual adaptation abilities, and fine-tuning them to downstream cross-lingual tasks has achieved promising results~\cite{pires2019multilingual, wu2019beto}.
Taking this further, better pre-trained language models have been proposed to improve the cross-lingual performance, such as using larger amounts of pre-trained data with larger pre-trained models~\cite{conneau2019unsupervised,liang2020xglue}, and utilizing more tasks in the pre-training stage~\cite{huang2019unicoder}.

\begin{figure}[!t]
\begin{subfigure}{0.45\textwidth}
    \centering
    \includegraphics[scale=0.445]{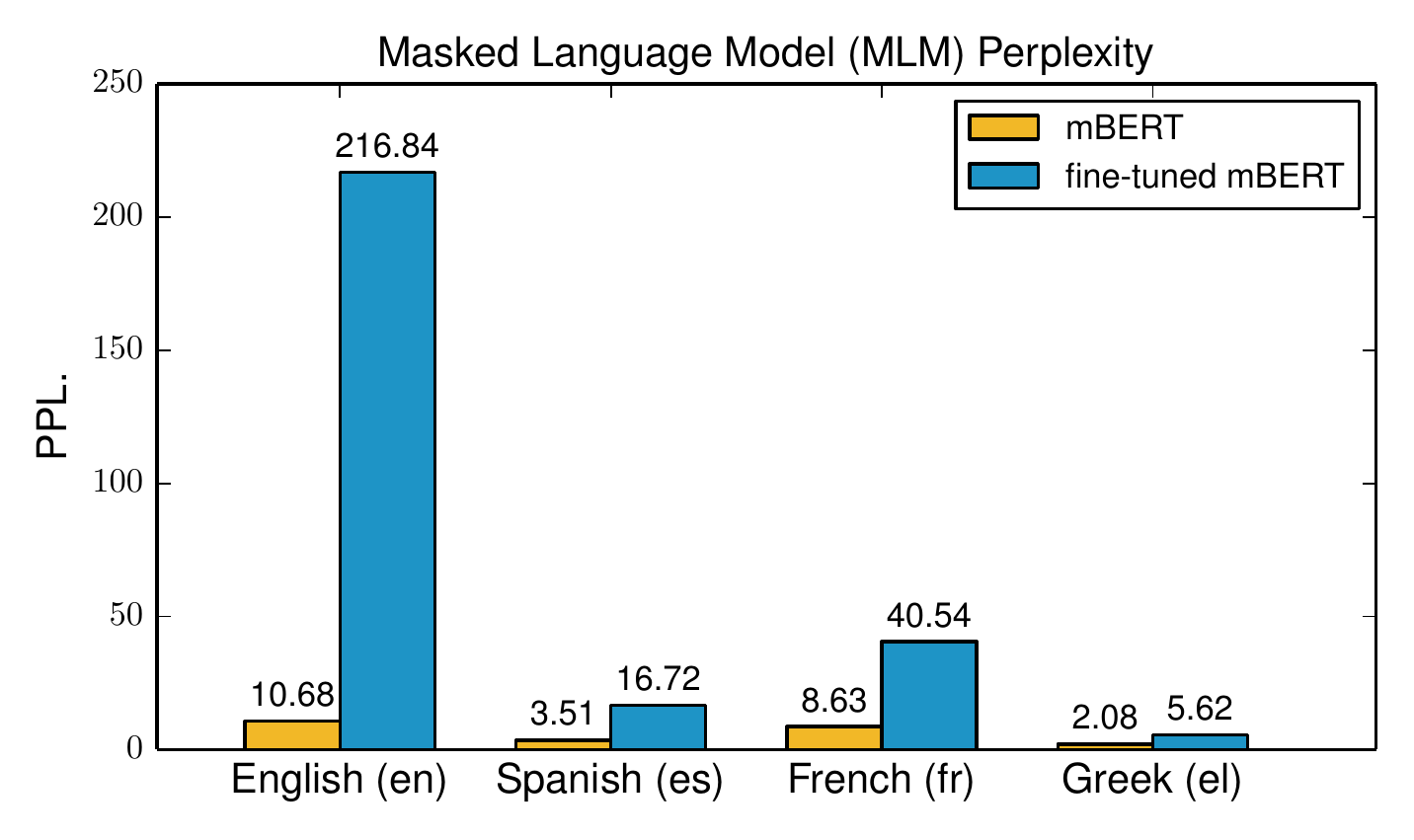}
\end{subfigure}
\begin{subfigure}{0.45\textwidth}
    \centering
    \includegraphics[scale=0.445]{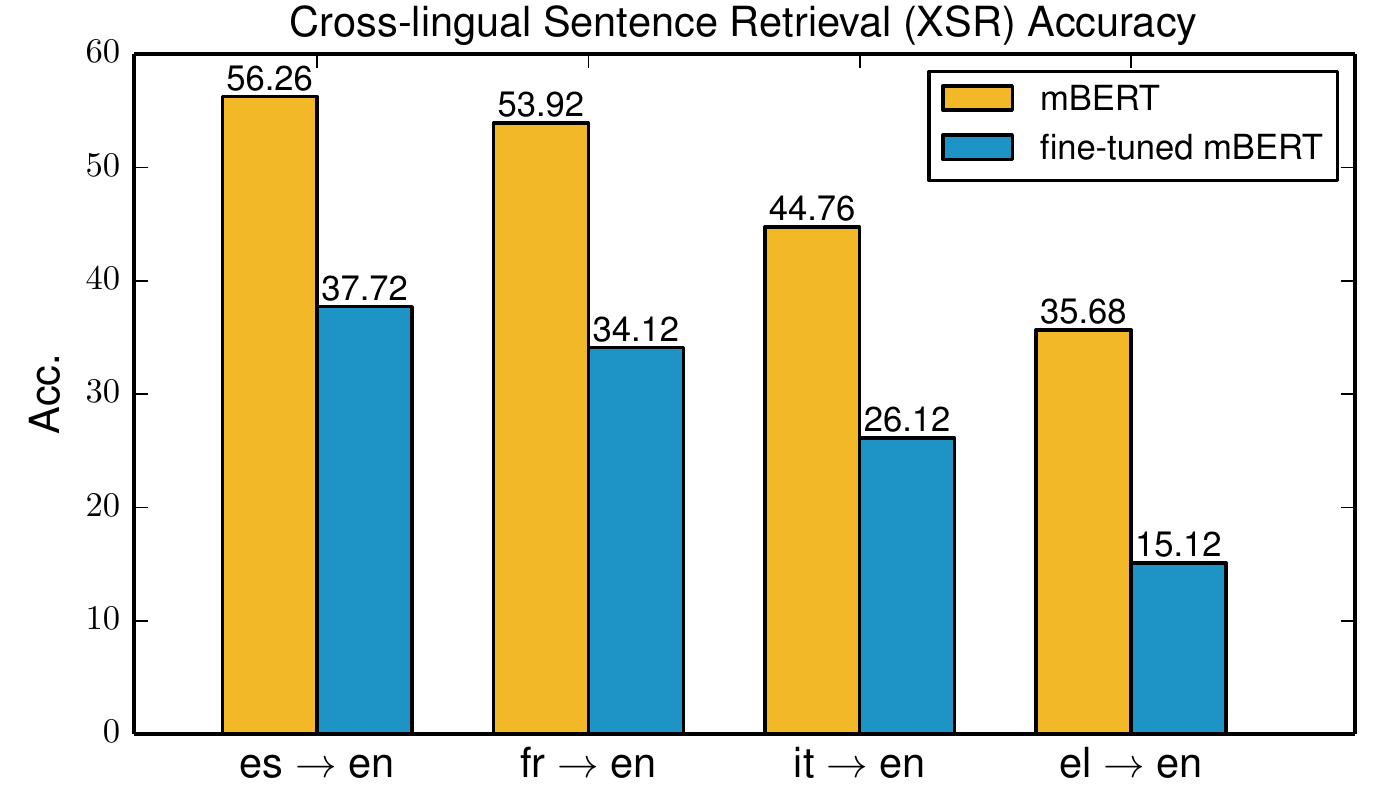}
\end{subfigure}
\caption{Masked language model and cross-lingual sentence retrieval results before and after fine-tuning mBERT to the English part-of-speech tagging task.}
\label{fig:mlm_xsr}
\end{figure}

However, we observe that multilingual BERT (mBERT)~\cite{devlin2019bert}, a pre-trained language model, forgets the masked language model (MLM) task that has been learned and partially loses the cross-lingual ability (from a cross-lingual sentence retrieval (XSR)\footnote{This task is to find the correct translation sentence from the target corpus given a source language sentence.} experiment) after being fine-tuned to the downstream task in English, as shown in Figure~\ref{fig:mlm_xsr}, which results in sub-optimal cross-lingual performance to target languages.

In this paper, we consider a new direction to improve the cross-lingual performance, which is to preserve the cross-lingual ability of pre-trained multilingual models in the fine-tuning stage.
Motivated by the continual learning~\cite{ring1994continual,rebuffi2017icarl,kirkpatrick2017overcoming,lopez2017gradient} that aims to learn a new task without forgetting the previous learned tasks, we adopt a continual learning framework to constrain the parameter learning in the pre-trained multilingual model when we fine-tune it to downstream tasks in the source language. Specifically, based on the results in Figure~\ref{fig:mlm_xsr}, we aim to maintain the cross-linguality of pre-trained multilingual models by utilizing MLM and XSR tasks to constrain the parameter learning in the fine-tuning stage. 

Experiments show that our methods help pre-trained models better preserve the cross-lingual ability. Additionally, our methods surpass other fine-tuning baselines on the strong multilingual model mBERT and XLMR~\cite{conneau2019unsupervised} on zero-shot cross-lingual part-of-speech tagging (POS) and named entity recognition (NER) tasks.

\section{Related Work}
Cross-lingual methods, which alleviate the need for obtaining large amounts of annotated data in target languages, have been applied to multiple NLP tasks, such as task-oriented dialogue systems~\cite{chen2018xl,liu2019zero}, part-of-speech tagging~\cite{wisniewski2014cross,zhang2016ten,kim2017cross}, named entity recognition~\cite{mayhew2017cheap,ni2017weakly,xie2018neural}, abstractive summarization~\cite{duan2019zero,zhu2019ncls}, and dependency parsing~\cite{schuster2019cross,ahmad2019difficulties}.
Recently, multilingual language models~\cite{devlin2019bert,nipscross2019,huang2019unicoder,conneau2019unsupervised}, pre-trained on a large-scale data corpus across a great many languages, have significantly improved the cross-lingual performance.
However, the corresponding fine-tuning techniques have been less studied. To the best of our knowledge, only~\citet{wu2019beto} have investigated the effectiveness of fine-tuning mBERT by freezing its partial bottom layers.

\section{Methodology}
In this section, we first describe the gradient episodic memory (GEM)~\cite{lopez2017gradient}, a continual learning framework, which we adopt to constrain the fine-tuning process. 
Then, we introduce how we fine-tune the pre-trained multilingual model with GEM.

\subsection{Gradient Episodic Memory (GEM)}
We consider a scenario where the model has already learned $n-1$ tasks and needs to learn the $n$-th task.
The main feature of GEM is an episodic memory $\mathcal{M}_k$ that stores a subset of the observed examples from task $k$ ($k \in [1,n]$). The loss at the memories from the $k$-th task can be defined as
\begin{equation}
    \mathcal{L}(f_\theta, \mathcal{M}_k) = \frac{1}{|\mathcal{M}_k|} \sum_{(x_i,k,y_i)\in \mathcal{M}_k} \mathcal{L}(f_\theta (x_i,k), y_i),
    \label{eq1}
\end{equation}
where the model $f_\theta$ is parameterized by $\theta$.
In order to maintain the performance of the model in the previous $n-1$ tasks while learning the $n$-th task, GEM utilizes the losses for the previous $n-1$ tasks in Eq.~(\ref{eq1}) as inequality constraints, avoiding their increase but allowing their decrease. Concretely, when observing the training samples ($x$, $y$) from the $n$-th task, GEM solves the following problem:
\begin{align}
    & \text{minimize}_\theta  ~~ \mathcal{L}(f_\theta (x,n),y) \nonumber \\
    & \text{subject to} \nonumber \\
    & \mathcal{L}(f_\theta, \mathcal{M}_k) \leq \mathcal{L} (f_\theta^{n-1}, \mathcal{M}_k) ~ \text{for all} ~ k < n, \label{eq2}
\end{align}
where $f_\theta^{n-1}$ is the model before learning task $n$.

\subsection{Fine-tuning with GEM}
We consider two tasks ($n=2$) in total by applying GEM to the fine-tuning of pre-trained multilingual models, namely, mBERT and XLMR. The first task is either what the pre-trained models have already learned (MLM) or the ability that they already possess (XSR), and the second task is the fine-tuning task.
We follow Eq.~(\ref{eq2}) when we fine-tune the pre-trained models:
\begin{align}
    & \text{minimize}_\theta  ~~ \mathcal{L}(f_\theta (x,\mathcal{T}_2), y) \nonumber \\
    & \text{subject to} ~~ \mathcal{L}(f_\theta, \mathcal{T}_1) \leq \mathcal{L} (f_\theta^*, \mathcal{T}_1),
\end{align}
where $\mathcal{T}_1$ and $\mathcal{T}_2$ denote the first and second tasks, respectively, and $f_\theta^*$ represents the original pre-trained model. 
When the MLM task is considered as the first task, we constrain the fine-tuning process of the pre-trained model by preventing it from forgetting its original task after fine-tuning so as to better preserve the original cross-lingual ability. When the XSR task is considered as the first task, on the other hand, we prevent the pre-trained model from losing its cross-lingual ability after fine-tuning. We also consider incorporating both MLM and XSR as the first task.

\begin{table*}[!t]
\centering
\resizebox{0.97\textwidth}{!}{
\begin{tabular}{lcccccCCCCCC}
\hline
\multicolumn{1}{l|}{\multirow{2}{*}{\textbf{Model}}} & \multicolumn{5}{c|}{\textbf{MLM}} & \multicolumn{3}{c|}{\textbf{XSR (Spanish to English)}} & \multicolumn{3}{c}{\textbf{XSR (Italian to English)}} \\ \cline{2-12}
\multicolumn{1}{l|}{} & \multicolumn{1}{c}{\multirow{1}{*}{$
\tt{en}$}} & \multicolumn{1}{c}{\multirow{1}{*}{$\tt{es}$}} & \multicolumn{1}{c}{\multirow{1}{*}{$\tt{fr}$}} & \multicolumn{1}{c}{\multirow{1}{*}{$\tt{el}$}} & \multicolumn{1}{c|}{\multirow{1}{*}{$\tt{ru}$}} & \multicolumn{1}{C}{P@1} & \multicolumn{1}{C}{P@5} & \multicolumn{1}{C|}{P@10} & \multicolumn{1}{C}{P@1} & \multicolumn{1}{C}{P@5} & \multicolumn{1}{C}{P@10} \\ \hline
\multicolumn{1}{l|}{mBERT} & 10.68 & 3.51 & 8.63 & 2.08 & \multicolumn{1}{c|}{2.70} & 56.26 & 68.80 & \multicolumn{1}{C|}{73.92} & 44.76 & 61.32 & 66.70 \\ \hline
\multicolumn{1}{l|}{Naive Fine-tune} & 216.80 & 16.72 & 40.54 & 5.62 & \multicolumn{1}{c|}{8.61} & 37.72 & 52.20 & \multicolumn{1}{C|}{58.43} & 26.12 & 37.46 & 46.69 \\
\multicolumn{1}{l|}{\quad w/ frozen layers} & 95.17 & 9.33 & 30.04 & 3.44 & \multicolumn{1}{c|}{5.34} & 38.16 & 53.92 & \multicolumn{1}{C|}{59.16} & 28.69 & 42.74 & 48.76 \\ \hline
\multicolumn{12}{l}{\textbf{\textit{Multi-Task Learning}}} \\ \hline
\multicolumn{1}{l|}{MTF w/ MLM} & \underline{9.50} & \underline{5.10} & \underline{8.62} & \underline{2.56} & \multicolumn{1}{c|}{\underline{3.47}} & 35.93 & 50.41 & \multicolumn{1}{C|}{56.20} & 24.79 & 37.18 & 45.46 \\
\multicolumn{1}{l|}{MTF w/ XSR} & 121.50 & 100.10 & 96.50 & 773.00 & \multicolumn{1}{c|}{180.80} & 75.40 & 80.88 & \multicolumn{1}{C|}{85.76} & 75.94 & 85.44 & 88.29 \\
\multicolumn{1}{l|}{MTF w/ Both} & \underline{9.89} & \underline{9.45} & \underline{11.30} & \underline{3.80} & \multicolumn{1}{c|}{\underline{4.16}} & 77.84 & 82.57 & \multicolumn{1}{C|}{87.97} & 74.38 & 83.29 & 86.95 \\ \hline
\multicolumn{12}{l}{\textbf{\textit{Continual Learning}}} \\ \hline
\multicolumn{1}{l|}{GEM w/ MLM} & \underline{12.99} & \underline{6.62} & \underline{11.39} & \underline{2.87} & \multicolumn{1}{c|}{\underline{4.22}} & \textbf{42.90} & \textbf{57.26} & \multicolumn{1}{C|}{\textbf{63.58}} & \textbf{31.66} & \textbf{44.16} & \textbf{50.16} \\
\multicolumn{1}{l|}{GEM w/ XSR} & 252.9 & 26.73 & 55.95 & 11.84 & \multicolumn{1}{c|}{16.46} & 63.65 & 75.45 & \multicolumn{1}{C|}{80.56} & 63.56 & 78.18 & 83.42 \\
\multicolumn{1}{l|}{GEM w/ Both} & \underline{12.16} & \underline{6.40} & \underline{10.62} & \underline{3.40} & \multicolumn{1}{c|}{\underline{4.30}} & 64.34 & 76.23 & \multicolumn{1}{C|}{81.42} & 64.12 & 79.35 & 84.59 \\ \hline
\end{tabular}
}
\caption{Experiments on MLM and XSR tasks based on mBERT. Models other than mBERT are fine-tuned to the English POS task. The underlined numbers in the MLM task denote that the performance is close to mBERT's. The bold numbers in the XSR task denote the best performance after fine-tuning without using the XSR supervision.}
\label{table:mlm_and_xsr}
\end{table*}

\section{Experiments}

\subsection{Dataset}
For the POS task, we use Universal Dependencies 2.0~\cite{nivre2017universal} and select English (en), French (fr), Spanish (es), Greek (el) and Russian (ru) to evaluate our methods. 
For the NER task, we use CoNLL 2002~\cite{tjong2002introduction} and CoNLL 2003~\cite{sang2003introduction}, which contain English (en), German (de), Spanish (es) and Dutch (nl), to evaluate our methods. 
For both tasks, we consider English as the source language and other languages as target languages.

\subsection{Baselines}
We compare our methods to several baselines. \textbf{Naive Fine-tune}~\cite{wu2019beto} is to add one linear layer on top of the pre-trained model while fine-tuning with L2 regularization. \textbf{Fine-tune with Partial Layers Frozen} \cite{wu2019beto} is to fine-tune pre-trained multilingual models by freezing the partial bottom layers. And \textbf{Multi-Task Fine-tune (MTF)} is to fine-tune pre-trained multilingual models on both the fine-tuning task and additional tasks (MLM and XSR).\footnote{The code is attached in the supplementary material.}




\subsection{Training Details}
We conduct the MLM task with two settings. First, we only utilize the English Wikipedia corpus (\textbf{MLM (en)}) since we observe the catastrophic forgetting in the English MLM task as in Figure~\ref{fig:mlm_xsr}. Second, we utilize both the source and target languages Wikipedia corpus (\textbf{MLM (all)}). The first setting is used in our main experiments.
Note that we do not use all pre-trained languages in mBERT for the MLM task because it would make the fine-tuning process very time-consuming.
For the XSR task, we leverage the sentence pairs between the source and target languages from the Europarl parallel corpus~\cite{koehn2005europarl}.\footnote{More training details are in the appendix.}


\section{Results \& Analysis}


\paragraph{Does GEM preserve the cross-lingual ability?}
From Table~\ref{table:mlm_and_xsr}, we can see that naive fine-tuning mBERT significantly decreases the MLM performance, especially in English. Since mBERT is fine-tuned to the English task, the English subword embeddings are fine-tuned, which makes mBERT lose more MLM task information in English. Naive fine-tuning also makes the XSR performance of mBERT drop significantly.
We observe that fine-tuning with partial layers frozen is able to somewhat prevent the MLM performance from getting worse, while fine-tuning with GEM based on that task almost preserves the original MLM performance of mBERT. Although we only use English data in the MLM task, using GEM based on the MLM task still preserves the task-related parameters that are useful for other languages. Correspondingly, we can see that \textit{GEM w/ MLM} achieves better XSR performance than \textit{Naive Fine-tune w/ frozen layers}, which shows that GEM helps better preserve the cross-lingual ability of mBERT.

In addition, although \textit{GEM w/ XSR} aggravates the catastrophic forgetting in the MLM task, it is able to significantly improve the XSR performance due to the usage of the XSR supervision. Furthermore, incorporating both the MLM and XSR tasks can better preserve the performance in both tasks. 

\begin{table*}[!t]
\centering
\resizebox{0.945\textwidth}{!}{
\begin{tabular}{lccccc|l|cccc|l}
\hline
\multicolumn{1}{l|}{\multirow{2}{*}{\textbf{Model}}} & \multicolumn{6}{c|}{\textbf{POS}} & \multicolumn{5}{c}{\textbf{NER}} \\ \cline{2-12}
\multicolumn{1}{l|}{} & \multicolumn{1}{c}{$\tt{en}$} & \multicolumn{1}{c}{$\tt{es}$} & \multicolumn{1}{c}{$\tt{fr}$} & \multicolumn{1}{c}{$\tt{el}$} & \multicolumn{1}{c|}{$\tt{ru}$} &
\multicolumn{1}{c|}{$\tt{avg}^{\dagger}$} &
\multicolumn{1}{c}{$\tt{en}$} & \multicolumn{1}{c}{$\tt{es}$} & \multicolumn{1}{c}{$\tt{de}$} & \multicolumn{1}{c|}{$\tt{nl}$} & \multicolumn{1}{c}{$\tt{avg}^{\dagger}$} \\ \hline
\multicolumn{1}{l|}{Naive Fine-tune} & 96.23 & 82.95 & 89.12 & 84.21 & 85.45 & 85.43 & \textbf{91.97} & 74.96 & 69.56 & 77.57 & 74.03 \\
\multicolumn{1}{l|}{\quad w/ frozen layers} & 96.07 & 83.41 & 89.41 & 85.54 & 85.17 & 85.88 & 91.90 & 75.27 & 70.23 & 77.89 & 74.46 \\ \hline
\multicolumn{11}{l}{\textbf{\textit{Multi-Task Learning}}} \\ \hline
\multicolumn{1}{l|}{MTF w/ MLM} & 94.47 & 83.01 & 88.08 & 84.48 & 80.46 & 84.01 & 91.82 & 71.47 & 67.90 & 74.91 & 71.43 \\
\multicolumn{1}{l|}{MTF w/ XSR} & 96.39 & 82.41 & 87.05 & 72.51 & 86.09 & 82.01 & 91.85 & 74.02 & 68.55 & 75.67 & 72.75 \\ 
\multicolumn{1}{l|}{MTF w/ Both} & 95.63 & 83.52 & 89.07 & 85.21 & 83.10 & 85.28 & 91.74 & 71.87 & 68.12 & 74.86 & 71.62 \\ \hline
\multicolumn{1}{l}{\textbf{\textit{Continual Learning}}} \\ \hline
\multicolumn{1}{l|}{GEM w/ MLM} & \textbf{97.39} & 84.65 & 89.74 & 86.04 & \textbf{86.93} & 86.84$^{\ddagger}$ & 91.93 & \textbf{76.45} & 70.48 & 78.61 & 75.18$^{\ddagger}$ \\
\multicolumn{1}{l|}{GEM w/ XSR} & 96.97 & 84.53 & 89.83 & \textbf{86.53} & 86.36 & \multicolumn{1}{c|}{86.81$^{\ddagger}$} & 91.89 & 76.29 & 70.74 & 78.77 & 75.27$^{\ddagger}$ \\ 
\multicolumn{1}{l|}{GEM w/ Both} & 97.04 & \textbf{84.91} & \textbf{90.32} & 86.44 & 86.13 & \multicolumn{1}{c|}{\textbf{86.95}$^{\ddagger}$} & 91.45 & 76.20 & \textbf{70.98} & \textbf{79.19} & \textbf{75.46}$^{\ddagger}$ \\\hline
\end{tabular}
}
\caption{Zero-shot results on POS and NER tasks based on mBERT. $^{\dagger}$The average scores excluding $\tt{en}$. $^{\ddagger}$The results are statistically significant compared to all baselines with $p < 0.01$ by t-test.}
\label{table:pos_and_ner}
\end{table*}

\paragraph{Does GEM improve the cross-lingual performance?}
From Table~\ref{table:pos_and_ner}, we can see that our methods consistently surpass the fine-tuning baselines on all target languages in the POS and NER tasks. In terms of the average performance, our methods outperform the baselines by an around or more than 1\% improvement.\footnote{The results of XLMR are included in the appendix due to the page limit. It will be added to the final version.}
In addition, constraining mBERT fine-tuning on the MLM task shows similar performance to constraining it on the XSR task. We conjecture that the effectiveness of both methods is similar, although they come from different angles. When the information of both tasks is utilized, GEM is able to slightly improve the performance. We find that the experimental results on XLMR are consistent with mBERT.

\begin{table}[!t]
\centering
\resizebox{0.49\textwidth}{!}{
\begin{tabular}{l|l|ccccc|c}
\hline
\multicolumn{1}{c|}{\textbf{Task}}                 & \multicolumn{1}{c|}{\textbf{Models}}            & $\tt{en}$   & $\tt{es}$   & $\tt{fr}$   & $\tt{el}$   & $\tt{ru}$   & $\tt{avg}$  \\ \hline 
\multirow{5}{*}{MLM} & mBERT             & 10.7 & \textbf{3.51} & 8.63 & \textbf{2.08} & \textbf{2.70} & 5.52 \\
                     & MTF w/ MLM (en)  & 9.50 & 5.10 & 8.62 & 2.56 & 3.47 & 5.85 \\
                     & MTF w/ MLM (all) & \textbf{9.33} & 4.19 & \textbf{4.89} & 2.34 & 3.04 & \textbf{4.76} \\
                     & GEM w/ MLM (en)   & 13.0 & 6.62 & 11.4 & 2.87 & 4.22 & 7.62 \\
                     & GEM w/ MLM (all)  & 11.8 & 4.18 & 6.83 & 2.29 & 2.99 & 5.62 \\ \hline
\multirow{5}{*}{POS} & Naive Fine-tune     & 96.2 & 82.9 & 89.1 & 84.2 & 85.5 & 85.4 \\
                     & MTF w/ MLM (en)  & 94.5 & 83.0 & 88.1 & 84.5 & 80.5 & 84.0 \\
                     & MTF w/ MLM (all) & 94.7 & 77.5 & 83.3 & 81.9 & 77.0 & 79.9 \\
                     & GEM w/ MLM (en)   & 97.4 & \textbf{84.7} & \textbf{89.7} & \textbf{86.0} & 86.9 & \textbf{86.8} \\
                     & GEM w/ MLM (all)  & 97.2 & 83.9 & 89.2 & 85.9 & \textbf{87.1} & 86.5 \\ \hline
\end{tabular}
}
\caption{Ablation study on the two settings of using the MLM task based on mBERT.
}
\label{ablation_study}
\end{table}


\paragraph{GEM vs. MTF}
From Table~\ref{table:mlm_and_xsr}, we notice that using the MLM task, MTF achieves lower perplexity than GEM since it aggressively trains mBERT on this task. However, we observe that \textit{MTF w/ MLM} makes the performance of the XSR, POS and NER tasks worse than \textit{Naive Fine-tune}, and we speculate that MTF pushes mBERT to be overfit on the MLM task, instead of preserving its cross-lingual ability. Meanwhile, we can see that GEM regularizes the loss of the training on the MLM task to avoid catastrophic forgetting of previously trained languages, and conserve the cross-linguality of the pre-trained multilingual models. 

In addition, we observe that adding XSR objective to the training cause the MLM performance worse.
Although MTF achieves the best performance in the XSR task since it directly fine-tunes mBERT on that task, we can see from Table~\ref{table:pos_and_ner} that \textit{GEM w/ XSR} boosts the cross-lingual performance of downstream tasks, while \textit{MTF w/ XSR} has the opposite effect.
We speculate that brutally fine-tuning mBERT on the XSR task (\textit{MTF w/ XSR}) just makes mBERT learn the XSR task, while using GEM to constrain the fine-tuning on the XSR task can preserve its cross-lingual ability of mBERT. Incorporating both the MLM and XSR tasks further improves the performance for GEM, while MTF still performs worse than \textit{Naive Fine-tune}.



\paragraph{Ablation Study}
From Table~\ref{ablation_study}, we can see that using GEM to constrain fine-tuning on MLM with all languages (\textit{GEM w/ MLM (all)}) achieves better performance than it does with only English (\textit{GEM w/ MLM (en)}) on the MLM task since more MLM supervision signals are provided, while their performances in the POS task are similar. Intuitively, since \textit{GEM w/ MLM} is able to improve the cross-lingual performance, constraining on more languages should give better performance. 
We conjecture, however, that the constraint with all languages could be too aggressive, so mBERT might tend to be overfit to the monolingual MLM task in all languages instead of preserving its original cross-lingual ability.
In addition, we observe that fine-tuning mBERT on the MLM task (MTF) would get worse when more languages are utilized.


\section{Conclusion}
In this paper, we propose to preserve the cross-linguality of pre-trained language models in the fine-tuning stage. To do so, we adopt a continual learning framework, GEM, to constrain the parameter learning in pre-trained multilingual models based on the MLM and XSR tasks when we fine-tune them to downstream tasks.
Experiments on the MLM and XSR tasks illustrate that our methods can better preserve the cross-lingual ability of pre-trained models. Furthermore, our methods achieve better performance than fine-tuning baselines for the strong multilingual models mBERT and XLMR on the zero-shot cross-lingual POS and NER tasks.

\bibliography{emnlp2020}
\bibliographystyle{acl_natbib}

\end{document}